\documentclass{article}
\usepackage[utf8]{inputenc}
\usepackage{amsmath}
\usepackage{hyperref}
\usepackage{graphicx}
\usepackage{float}
\usepackage[T1]{fontenc}
\usepackage[french]{babel}
\usepackage{csquotes}
\usepackage[utf8]{inputenc}
\usepackage[table]{xcolor}
\usepackage{caption}
\usepackage{array}        
\usepackage{multirow}     
\usepackage{colortbl}     
\usepackage{array}
\usepackage{etoolbox}
\usepackage{authblk}
\AtBeginEnvironment{tabular}{\centering}
\setkeys{Gin}{keepaspectratio=true}
\usepackage{enumitem}
\setlist[itemize]{label=\textbullet}

\usepackage[T1]{fontenc} 
\usepackage{lmodern} 
\usepackage{microtype} 

\usepackage[shortcuts]{extdash} 
\usepackage{hyphenat} 

\addto\captionsfrench{}
\addto\captionsfrench{\renewcommand{\abstractname}{Abstract}}
\title{\textbf{Synthetic Categorical Restructuring \\ 
\large Or How AIs Gradually Extract Efficient Regularities from Their Experience of the World}}

\author{Michael Pichat$^{1,2,4}$, William Pogrund$^{1,5}$, Paloma Pichat$^{1,3}$, Armanouche Gasparian$^{1}$, Samuel Demarchi$^{1,4}$, Martin Corbet$^{1,2}$, Alois Georgeon$^{1,2}$,Théo Dasilva$^{1,2}$, Michael Veillet-Guillem}

\date{}

\affil[1]{Neocognition (Chrysippe R\&D)}
\affil[2]{Facultés Libres de Philosophie et de Psychologie de Paris (ER IPC)}
\affil[3]{Faculté de Médecine de Lyon Est (Université Lyon 1)}
\affil[4]{Université Paris 8}
\affil[5]{INP-PHELMA, Université Grenoble Alpes}
\begin{document}

\maketitle

\renewcommand{\abstractname}{Abstract}
\begin{abstract}
How do language models segment their internal experience of the world of words to progressively learn to interact with it more efficiently? This study in the neuropsychology of artificial intelligence investigates the phenomenon of synthetic categorical restructuring—a process through which each successive perceptron neural layer abstracts and combines relevant categorical sub-dimensions from the thought categories of its previous layer. This process shapes new, even more efficient categories for analyzing and processing the synthetic system’s own experience of the linguistic external world to which it is exposed. Our \href{https://neuron-viewer.neocognition.org/1/1}{genetic neuron viewer}, associated with this study, allows visualization of the synthetic categorical restructuring phenomenon occurring during the transition from perceptron layer 0 to 1 in GPT2-XL.

\end{abstract}

\section{Context of Our Study}

\subsection{Mathematico-Cognitive Factors of Synthetic Categorical Segmentation}

In a previous study \cite{Pichat2024d}, we examined the mathematico-cognitive elements influencing the categorical segmentation performed by artificial neural networks in language models. In this exploratory analysis, we quantitatively and qualitatively investigated the factors that genetically influence this synthetic partitioning. Based on the aggregation function, expressed as $\sum(w_{i,j} x_{i,j}) + b$, which plays a central role in this cognitive process, we identified three factors, both mathematical and cognitive, that contribute to this conceptual compartmentalization.

First, the \textit{x-effect}, or synthetic categorical priming, refers to the fact that the activation of synthetic thought categories represented by neurons in layer $n$ influences the activation of categories in neurons strongly connected to them in layer $n+1$. In other words, the more a token belongs to an initial category in layer $n$ (i.e., the more it activates its associated neuron), the more likely it is to belong to categories in layer $n+1$ that are strongly linked to this precursor category. This priming phenomenon of prior categories thus guides, to some extent, the shaping of their superordinate categories (with which they have a strong connection weight) in layer $n+1$ by contributing to the determination of tokens that become part of the extension of the categories represented by neurons in this new layer.

Second, the \textit{w-effect}, or synthetic categorical attention, concerns the fact that the strength of connection weights between a target neuron (layer $n+1$) and its precursor neurons (layer $n$) influences the importance given to their associated initial categories in the formation of the category of this target neuron. This manifests qualitatively as a process of categorical complementation, genetically consisting of "bringing" to the extension (of tokens) of an arriving category a specific and distinct categorical sub-dimension extracted from each precursor category. This contribution is a function of the intensity of attentional focus each precursor category receives. The category of a target neuron is thus constituted, segment by segment, by semantically complementary categorical antecedents extracted from the extension of its subcategories.

Finally, the \textit{$\sum$-effect}, or synthetic categorical phasing, occurs when the same tokens are simultaneously activated within different precursor categories that are significantly associated with the same target neuron. This results in categorical resonance, which partially determines the tokens forming the extension of the category of this target neuron. From a qualitative perspective, this process manifests as a phenomenon of categorical intersection, genetically defining the content (in terms of tokens) of the categorical extension of this target category. Here, the extraction of categorical sub-dimensions from the precursor categories involves selecting sub-dimensions that are partially common among these precursor categories, rather than extracting distinct and complementary sub-dimensions within each precursor category, as seen in the previous \textit{w-effect}.

These three mathematico-cognitive causal factors of categorical segmentation drive, at the level of a target neuron (layer $n+1$), a process of extracting specific categorical sub-dimensions from the categories carried by its precursor neurons (layer $n$). These extracted precursor categorical sub-dimensions, assembled by the aggregation function at the level of the target neuron, thus genetically define the content of the category of this superordinate neuron. This synthetic concept extraction process, which has been extensively examined in humans \cite{Bolognesi2020, Haslam2020, EysenckKean2020, Fel2024, BathiaRichie2024, Marconato2024, Zettersten2024}, is a fascinating epistemological subject and contributes to the construction of "reality" as perceived by artificial cognition in its interaction with the world of tokens presented to it.

\subsection{Synthetic Categorical Clipping}

In another preliminary study \cite{Pichat2024e}, we highlighted that these three mathematico-cognitive factors of categorical segmentation govern a process of synthetic categorical clipping—an operation that consists of shaping and distinguishing a form from a categorical background. The objective was thus to understand the properties of this categorical clipping, which operates on the relative categorical variability of the tokens constituting the extension of each precursor neuron's category to extract a subset of categorically homogeneous tokens aligned with the (new) specific category created and carried by their corresponding target neuron.

In our exploratory analysis, we identified several synthetic characteristics of this categorical clipping:

\begin{itemize}
    \item \textbf{Categorical reduction}, which reflects the fact that the categorical sub-dimension extracted from a precursor neuron's category groups semantically more homogeneous tokens compared to the tokens constituting the extension of the initial category—at least from the observational semantic reference frame constituted by the initial embeddings of GPT2-XL.
    \item \textbf{Categorical selectivity}, as a corollary, referring to the extraction of a significantly reduced set of tokens from the entire set of tokens forming the extension of each antecedent neuron's category.
    \item \textbf{Separation of initial embedding dimensions}, related to the fact that the clipping applied to a source category, when observed in the vector space of GPT2-XL embeddings, tends to manifest as a compartmentalization of these embeddings—some being preferentially paired with the extracted categorical form (i.e., the sub-dimension), while others remain more associated with the remaining categorical background of the initial category.
    \item \textbf{Segmentation of categorical zones in reduced embedding dimensions (down to two dimensions)}, observable as a relative distancing of the categorical centers of gravity of the extracted form and the non-retained background, each barycenter positioned in distinct categorical regions.
\end{itemize}

These elements provide insight into various synthetic cognitive properties through which categorical clipping creates extractions of categorical sub-dimensions from precursor neurons, grouping tokens into a form that converges as a fabricated homogeneous categorical segment.

The two preliminary studies we have summarized propose a series of explanatory notions that help us conceptualize how synthetic cognition extracts, from layer $n$, and during the transition from this layer to layer $n+1$, singular categorical sub-dimensions that will constitute the specific category of each target neuron. However, how can we further understand and describe the specific synthetic cognitive modalities through which the mathematico-cognitive factors of categorical segmentation (priming, attention, and phasing) and the resulting categorical clipping shape the categorical restructuring that occurs during the transition from one neural layer to the next? How, in more detail, do the mentioned synthetic cognitive processes (categorical priming, attention, phasing, and clipping) operate and combine to generate, at each new layer, a reorganization of the categorical segmentation system carried by the neurons of this new layer? What relationships do the newly reorganized superordinate categories maintain with their genetically subordinate categories?

\section{Epistemological and Conceptual Reflections on the Notion of Synthetic Categorical Restructuring}

\subsection{The Structuring of Thought Categories in Human Cognitive Psychology}

In the field of human cognition, an initial approach to the structuring of thought categories is proposed through the paradigm of semantic network representations \cite{CollinsQuillian1970, Sartori2024, Yang2024}. Here, concepts are represented and stored in long-term memory as semantic nodes, each node denoting a concept. Through associative links, each concept is connected to multiple other concepts, each expressing a characteristic of the given concept. Categories are then organized hierarchically according to their level of generality: superordinate concepts are spatially situated above subordinate ones. This follows a principle of cognitive economy: information stored at level $n$ is not redundantly stored at another superior or inferior level, ensuring that memory conservation occurs at the highest level of generality. When analyzing a given object, the retrieval of stored information occurs via a process of spreading activation \cite{Anderson1983}: the activation of a concept propagates to concepts semantically related to it within the categorical structure. However, this modeling of categorical structuring has been subject to partial empirical contradictions, notably concerning issues of semantic resemblance \cite{Conrad1972} and typicality \cite{Rosch1978}.

Research on categorical structuring through concept comparison \cite{Smith1974, Polyn2024, Planchuelo2024} offers a process-oriented alternative. This approach posits that when evaluating the relationship between two concepts, the first phase involves encoding the entities associated with each concept (e.g., bird and animal). Then, their respective semantic features are retrieved, distinguishing between defining traits (essential and constitutive of the category) and characteristic traits (frequent and common but non-essential). A proximity assessment follows, based on a similarity index calculated from the two sets of retrieved traits. If the computed value remains ambiguous (in terms of categorical proximity vs. difference), the comparison is restricted to defining traits only, enabling a final decision.

In a more refined yet complex approach, other studies on categorical structuring in memory \cite{Tversky1977, Zhang2024, Giallanza2024} suggest that the relationship between two categories is computed by integrating both common and divergent traits. The process involves retrieving each concept’s features from memory, determining the number of shared characteristics, identifying the number of differing traits, assigning weighted values to common versus distinct properties, and ultimately subtracting these weighted values to establish a proximity index.

From a rather contrasting perspective, studies on propositional representations \cite{BrewerHay1984, Gernsbacher1985, Yang2024, Moreira2024} assert that the fundamental meaningful unit is the proposition, which organizes categories through interrelations. These propositions, stored and retrieved in long-term memory, encode relationships between concepts in the form of more or less elaborate combinations of predicates and arguments with true or false truth values. This theoretical stance closely aligns epistemologically with the paradigm of ontologies in NLP and, more broadly, in machine learning \cite{PonomarevAgafonov2022, Patel2023, Thukral2023, Qiu2023, OrtizRodriguez2024}.

Finally, categorical structuring has also been modeled through schematic representations \cite{BowerBlackTurner1979, Hastie2022}. In this theoretical framework, broad-spectrum representations are stored in memory concerning world-related concepts and events. These are synthesized into general schemas, whose scripts (organized along a temporal dimension) represent their most frequently encountered instances. The function of these schemas is, for example, to store only the major steps of a given concept or knowledge sequence in memory, allowing for cognitive economy.

The various perspectives we have just synthesized, within the field of human cognitive psychology, on the possible modalities of category structuring allow us to fully grasp—if it were still necessary—the difference between human and artificial categorical cognition. Indeed, these different models are poorly suited to the specificities of synthetic categorical cognition. This is the case for at least the following three reasons:

\begin{itemize}
    \item These models are \textbf{static}: They aim to represent the fixed structuring of thought categories, whereas synthetic neural categorical systems are fundamentally genetic (albeit synchronic rather than diachronic): neural layers are embedded systems progressively constituting increasingly elaborate categories.
    \item These models are based on a \textbf{principle of semantic uniqueness} in an Aristotelian logic of the excluded middle, whereas synthetic categories exhibit significant polysemy \cite{Bills2023, Bricken2023}, at least when examined within a human semantic reference framework.
    \item These models \textbf{do not account for the elective categorical extraction performed by neurons}: The category of each neuron in layer $n+1$ does not simply integrate all of its subordinate categories; instead, it selectively extracts particular sub-dimensions.
\end{itemize}

Given that the paradigms of category structuring in human cognitive psychology have a static rather than dynamic purpose, they are not sufficiently relevant for understanding and describing the synthetic cognitive modalities through which mathematico-cognitive factors of categorical segmentation and categorical clipping generate the categorical restructuring that occurs during the transition from one neural layer to its superordinate layer.

\subsection{Extraction of Categorical Regularities and Categorical Restructuring}

What psychological conceptual framework should be mobilized to heuristically analyze, under appropriate transposition, the combination of synthetic cognitive processes we previously discussed (categorical priming, attention, phasing, and clipping) in the generation, at each new synthetic layer, of the restructuring of the categorical segmentation system carried by this new layer?

At this stage, an epistemological reflection seems necessary, which we will position within a constructivist framework. This approach appears more relevant for understanding the nature of cognitive activity performed by artificial neural systems in relation to the categorical restructuring these networks undertake. This epistemological reflection begins with a human psychological framework and attempts to transpose it, in an adjusted manner, to the realm of synthetic cognition.

Synthetic thought, like human cognition, does not produce internal representations that are mere analogical copies—passive recordings of the world’s properties, assumed to be pre-given and preexisting independently of observation. Indeed, knowledge is not a mirror of nature \cite{Varela1988}. This stands in contrast to empiricist or realist epistemologies, which assume a tendency "to think of knowledge as the representation of a world outside […] independent of the knower. The representation [is] supposed to reflect at least part of the world’s structure and the principles according to which it works" \cite[p.113]{vonGlaserfeld2002}.

Having established this interactionist position, how can we avoid falling into a solipsistic dead-end? Von Glaserfeld \cite[p.113]{vonGlaserfeld2002} clarifies: "if we were to say that there [is] no such relation [between knowledge and the objective world], we should find ourselves caught in solipsism, according to which the mind, and the mind alone, creates the world." The answer lies in the fact that intelligent activity is never passive but always serves a purpose \cite{Piaget1974}. This purpose is to adjust its contents and operational modalities to the unique and singular experience that the intelligent system constructs of the external world: "the cognizing subject has conceptually evolved in order to fit into the world as he or she experiences it" \cite[p.114]{vonGlaserfeld2002}.

But how does this adjustment occur? Primarily in a pragmatic rather than epistemic manner: "a living system, due to its circular organization, is an inductive system and functions always in a predictive manner: what happened once will occur again. Its organization (genetic and otherwise) is conservative and repeats only that which works" \cite[p.39]{Maturana1970}. For an intelligent system, this means identifying—or rather cognitively constructing—regularities \cite{Varela1984}, what Piaget \cite{Piaget1974} would call operational invariants, within the world as experienced through human sensory and cognitive frameworks, or within their synthetic correlates, the "techno-umwelts" \cite{Efimov2023}. As von Glaserfeld \cite[p.118]{vonGlaserfeld2002} states: "what we ordinarily call reality is the domain of the relatively durable perceptual and conceptual structures which we manage to establish, use, and maintain in the flow of our actual experience," adding: "empirical facts, from the constructivist perspective, are constructs based on regularities in a subject’s experience. They are viable if they maintain their usefulness and serve their purposes in the pursuit of goals" \cite[p.128]{vonGlaserfeld2002}.

In the specific context of synthetic cognition, what are these regularities, and what sustains them internally? The answer, at least within perceptron-based layers, is fairly immediate: these regularities are essentially the connection weights between neurons in successive layers (setting aside the issue of biases in the neural aggregation function). These weights are precisely what is learned during the deep learning phase of a neural network.

Moreover, these weights govern the specific activity of categorical restructuring during the construction of a new categorical neural layer from the previous one. These weights dictate synthetic attentional activity, which involves identifying, during the formation of a new artificial thought category in layer $n$, the precursor categories in layer $n-1$ that should be particularly focused on and to what degree (i.e., weighting). This aligns with von Glaserfeld's \cite[pp.90-91]{vonGlaserfeld2002} assertion in the domain of biological cognition: "focused attention picks a chunk of experience, isolates it from what came before and from what follows, and treats it as a closed entity," further emphasizing: "for this constructive activity, the role of attention is crucial […] Subjects can freely move their focus of attention in the perceptual field […] [attention is an] originator of coordination and relation" \cite[p.116]{vonGlaserfeld2002}.

The function of neural weights is fundamentally to compare and weight the level of attentional importance to be assigned to precursor synthetic categories in the differentiated coordination executed by the aggregation function during the genesis of their corresponding synthetic arrival categories. This is because the neural network has learned that such categorical restructuring is relevant for efficiently processing incoming information. As Humboldt \cite[p.581]{Humboldt1907} noted in the human cognitive domain: "in order to reflect, the mind must stand still for a moment in its progressive activity, must grasp as a unit what was just presented, and thus posit it as object against itself. The mind then compares the units […] and separates and connects them according to its needs."

At the conclusion of this section, attention appears as a potentially pertinent heuristic framework for cognitively analyzing the synthetic restructuring activity performed by neural networks.

\subsection{Attentional Processes and Categorical Restructuring}

Physiologically, attention arises from the nervous system’s limited capacity to process information and manifests through selective integration, activation, and use of sensory or memorized data (semantic, procedural) \cite{Funayama2024, BarrBieliauskas2024}. This results in an orientation reaction, focusing information retrieval on specific characteristics. In this context, neurocognitive studies of attentional processes—stemming from Posner’s approach \cite{Posner1995, Posner2024, Rueda2024}—highlight the role of a frontal attentional system within the frontal lobe (linked to conscious semantic focus and planning) and a posterior system within the parietal lobe (associated with visuospatial processes and shifting attentional locations).

In cognitive psychology, attention is defined as an adjustment of activity toward its goal, leading to increased efficiency in information collection (notably selectivity) and execution (in terms of precision and speed) for that specific activity \cite{PosnerSnyder1975, SchneiderShiffrin1977, Posner1978, Richard1980, TreismanGelade1980, Duncan1984, Tipper1985, Cowan2024, Wu2024, Gresch2024}. Regarding task execution, attention refers to the central nervous system’s control of activity, such as assigning importance (priority, order, reliability, etc.) to certain internal information (knowledge, representations, schemas) or supervising task performance quality over time.

Two cognitive functions are commonly assigned to attentional processes \cite{Duncan1999}:
\begin{itemize}
    \item \textbf{Signal detection}, primarily involving vigilance and exploration mechanisms to identify stimulus occurrences.
    \item \textbf{Selective attention}, enabling the focus on specific stimuli while disregarding others.
\end{itemize}

Vigilance is defined as the ability to selectively focus on a series of information over a given period to detect a targeted stimulus \cite{Mackworth1948}. This stimulus has a low occurrence rate but requires a rapid response \cite{Chen2024, Hanzal2024}. Vigilance is negatively impacted by uncertainty regarding the elements under attentional focus \cite{BroadbentGregory1965}. It can be modeled as an attentional spotlight \cite{Posner1980} and depends on expectations regarding the presence of the targeted element in a given location \cite{PosnerSnyderDavidson1980, Motter1999, Murray2024}.

Unlike vigilance, \textit{visual exploration} (or inspection) \cite{Zhang2024, Wu2024} is not a "passive" expectation of emerging targeted data but an active search for a stimulus \cite{PosnerDiGirolamo1998}. Typically, exploration is seen as a strategy for scanning specific features (traits, attributes) to locate them within an environment. In \textit{Feature Integration Theory} \cite{Treisman1986, Rosenholtz2024}, a mental map is assigned to each attribute, representing the occurrence of a specific feature in the visual field. These maps are examined in parallel, and the simultaneous investigation of a set of traits related to a single entity is made possible by an attentional process acting as a "mental glue," grouping involved attributes within the same representational area. This dynamic may involve inhibitory neuronal processes to filter out irrelevant characteristics. The \textit{Similarity Theory} \cite{DuncanHumphreys1992} conceptualizes attentional inspection in terms of proximity between target stimuli and non-relevant elements, as well as proximity among non-relevant elements themselves. Finally, the \textit{Guided Search Theory} \cite{CaveWolfe1990, Alahmari2024} envisions attentional exploration in two stages: (i) an initial phase activating a joint representation of all potential targets based on specific target characteristics, followed by (ii) a serial analysis of all potential targets' activation levels to select the most activated one.

Selective attention, which involves focusing on specific elements, is particularly studied through the \textit{cocktail party effect} \cite{Cherry1953, Liu2024, Bosker2024}, related to the elective tracking of a particular conversation among many. Three parameters influence attentional focus on the target speaker: (i) distinctive sensory properties, (ii) sound intensity, and (iii) spatial positioning. The \textit{Filter Theory of Attention} \cite{Broadbent1958, Zhao2024} suggests that sensory information from multiple sources reaches a filter that selects which will undergo further perceptual processing. However, this original model is increasingly replaced by the \textit{Attenuation Model} \cite{Treisman1964, Lin2024}, proposing that all incoming information is perceptually reduced in intensity, with only the least attenuated (those closest to targeted features) remaining. Finally, the \textit{Limited Attention Resource Model} \cite{Kahneman1973} addresses the parallel execution of activities concerning contrasting facilitation.

Selective attention is directly related to conceptualization processes \cite{Vergnaud2009, Vergnaud2016, Pichat2024c}. Indeed, when electively acquiring information, attention is linked to the creation of concepts, which entails identifying only relevant elements (associated with the objects of the activity) necessary for successful task completion. These elements help orient actions so they are well-adjusted and effective. Here, attention involves filtering and structuring an abundance of perceived information, excluding (or inhibiting) less relevant data to concentrate mental effort and selectivity on particular objects and characteristics previously demonstrated to be pragmatically significant.

This approach to attention, in a conceptualization framework, seems particularly well-suited to the notion of attentional weights (perceptron) in deep learning. Indeed, it involves learning to construct, at the perceptron neural layer of level \textit{n+1}, new categories of thought by combining, in a weighted and selective manner at the attentional level, the categories carried by the neurons of the previous \textit{n} layer. This conceptual approach is therefore the one we will particularly employ in our present study of the synthetic categorical restructuring process. This is all the more relevant as this conceptual approach is epistemologically coherent with the categorical investigation approach that underpins our work.

\section{Research Problem}

\subsection{Extraction of Attentional Regularities, Clipping Factors, and Categorical Restructuring}

Closely related to the notion of conceptualization, synthetic categorical attention \cite{Pichat2024d}, or the "w-effect," is a decisive mathematico-cognitive factor in categorical restructuring, the core subject of our present study. Indeed, neural attentional weights represent the learned regularities within the neural system that directly influence its selective attentional focus when forming a new thought category in layer $n$, based on specific categories in layer $n-1$. Consequently, these weights determine categorical clipping \cite{Pichat2024e}, which involves extracting a precise categorical sub-dimension \cite{Pichat2024b} from each precursor category to construct, through weighted combination, the newly reorganized superordinate thought category.

However, and this is the central objective of the present study, what are the specific cognitive modalities through which synthetic categorical attention supports the categorical restructuring carried out by neurons in layer $n$ based on the categories represented by neurons in the previous layer? How does the extraction of categorical sub-dimensions, enabled by categorical attention and constitutive of this restructuring, manifest in terms of activation and thus categorical priming? Through what mechanisms does categorical attention interact with categorical phasing to generate this restructuring? In other words, what are the phenomenological aspects through which these three mathematico-cognitive factors co-actively shape synthetic categorical clipping and restructuring?

Before presenting the methodological choices we made in our specific investigation of these questions, as well as the operational frameworks adopted, we take a moment below to further clarify what we mean by categorical restructuring and to define a series of methodological elements relevant to these operationalizations.

\subsection{Clarifications on the Phenomenon of Synthetic Categorical Restructuring}

Graph 1 illustrates a case of synthetic categorical restructuring, using the example of neuron 121 in perceptron layer 1 of GPT2-XL, in relation to three of its strongest precursor neurons in layer 0, with which it maintains the highest connection weights (and thus the highest attentional focus). It is important to clarify that the categorical clusters (subgroups of tokens) mentioned in relation to this target neuron and its precursors are not intrinsic semantic realities but rather the result of a semantic clustering operation performed by GPT-4o.

Let us begin with precursor neuron 372l in layer 0. By focusing on the 100 tokens that most strongly activate this neuron on average (which we term "core-tokens," as we define them as constituting the categorical extension of the implicated category), we can associate this neuron with the thought category "legal concepts," which this neuron is presumably detecting.

This category can be decomposed into five categorical clusters of tokens (within these 100 core-tokens):

\begin{itemize}
    \item \textbf{"Legal terms"}: including tokens such as \texttt{[patent], [liability], [arbitration]}, etc.
    \item \textbf{"Accusations"}: involving tokens such as \texttt{[allegation], [accusation], [disputed]}, etc.
    \item \textbf{"Claims \& charges"}: associated with tokens such as \texttt{[billing], [claim], [charge]}, etc.
    \item \textbf{"Legal actions"}: referring to tokens such as \texttt{[lawsuit], [sue], [embroiled], [litigation]}, etc.
    \item \textbf{"Legal roles"}: including tokens such as \texttt{[defendant], [lawyers], [plaintiff], [claimant]}, etc.
\end{itemize}

Among these 100 initial core-tokens, some also strongly activate target neuron 121 in layer 1 on average: \texttt{ [charges], [claims], [accusations], [allegations], [alleging]}, etc. These tokens thus also become core-tokens constituting the categorical extension of this target neuron. These specific tokens, which we term "taken-tokens," can therefore be interpreted as having been extracted and clipped from the precursor category to contribute to the formation of the target category. These taken-tokens form a "taken-cluster" (represented in green in the graph), which has been semantically interpreted as belonging to the categorical sub-dimension "legal allegations," extracted from the original category. Notably, this "legal allegations" sub-dimension is not simply a selection of one of the categorical clusters ("legal terms," "accusations," "claims \& charges," "legal actions," "legal roles") from the original "legal concepts" category. Instead, this "legal allegations" sub-dimension results from a genuine constructive categorical clipping process, selectively extracting and regrouping certain tokens initially distributed across different categorical clusters. This unique and original categorical sub-dimension thus clearly demonstrates categorical reconstruction—i.e., a new form of categorical segmentation of objects within the world of tokens.

\begin{figure}[H]
    \centering
    \hspace*{-.2\textwidth} 
    \includegraphics[width=1.4\textwidth]{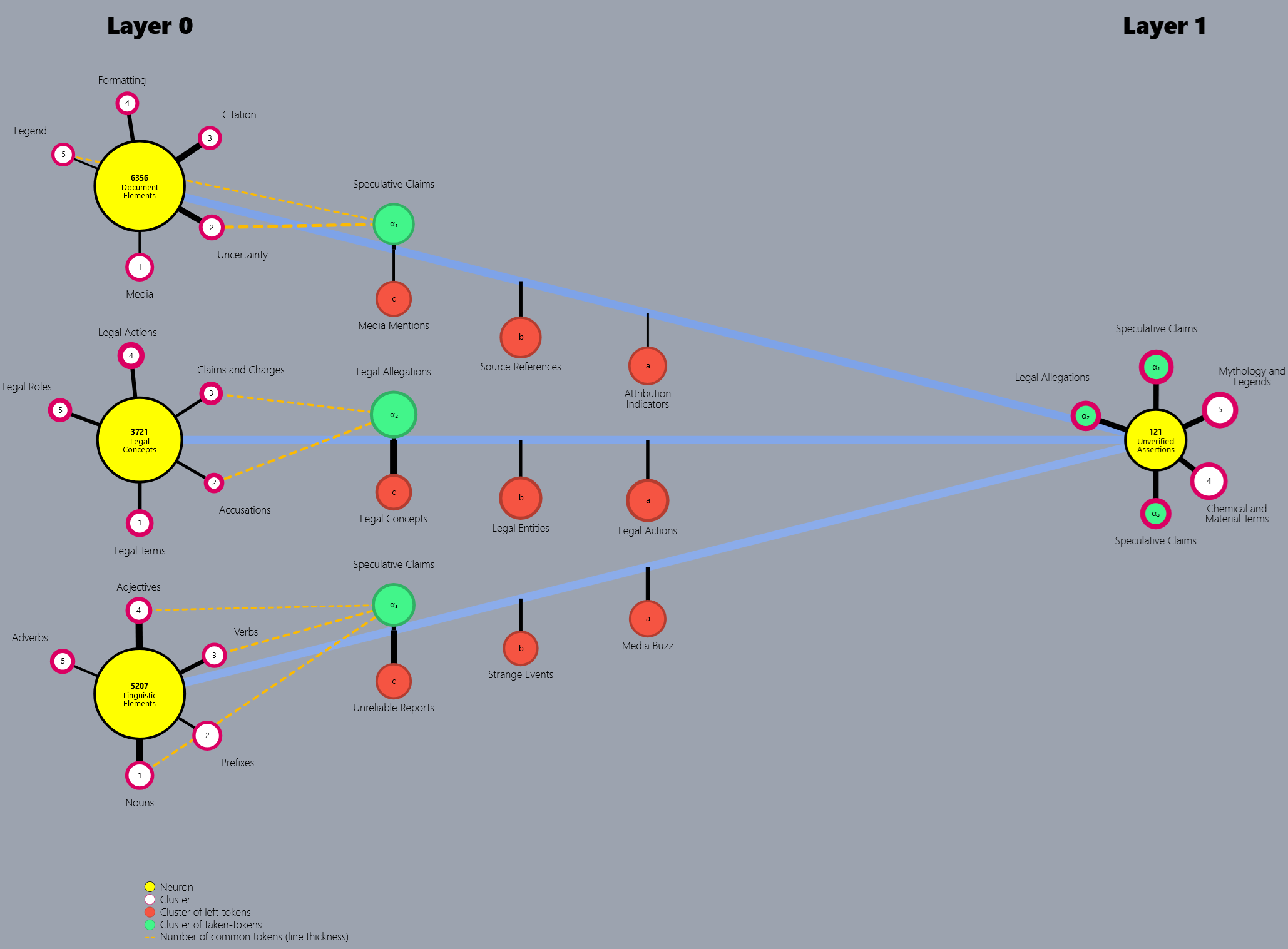}
    \captionsetup{justification=centering, font=small}
    \caption*{\textit{Graph n$^{\circ}$1: Example of categorical restructuring (case of neuron 121 in perceptron layer 1 of GPT2-XL, in relation to three of its precursor neurons with maximum connection weight).}}

    \label{fig:graph1}
\end{figure}

This synthetic mechanism of constructing and extracting a new categorical sub-dimension occurs for each precursor category. Each of these categorical sub-dimensions, in turn, becomes one of the categorical clusters of the target neuron, which in this case carries a category interpreted as belonging to "unverified assertions." Consequently, as precursor categories transition to their associated target category through categorical clipping, we observe an authentic categorical restructuring process carried out by the target neuron. This synthetic restructuring translates into the construction of the distinction: (i) of extracted categorical forms (the clipped categorical sub-dimensions, represented by the taken-clusters in green in the graph) and (ii), of a non-retained categorical background (represented by the "left-clusters" in red).

Our \href{https://neuron-viewer.neocognition.org/1/1}{genetic neuron viewer}, associated with this study, allows for the visualization of the synthetic categorical restructuring phenomenon occurring during the transition from perceptron layer 0 to 1 in GPT2-XL.

\subsection{Synthetic Partial Categorical Confluence}

A primary conceptual and methodological choice in operationalizing our investigation into synthetic categorical restructuring—specifically, how precursor categories in layer $n$ transition to their associated target category in layer $n+1$—is as follows.

What cognitive effect does the coactivity of synthetic categorical attention and categorical phasing have on artificial categorical restructuring during the transition from precursor categories to their associated category in a subsequent layer? To what extent does the interaction between these two mathematico-cognitive factors drive categorical clipping, which ultimately generates this restructuring? How does this manifest at the level of the clipped categorical sub-dimensions of each precursor category?

Here, we hypothesize the following effect: \textit{partial categorical confluence}, for a given target category in layer $n$, of its extracted genetic categorical sub-dimensions from layer $n-1$. In other words, a relative semantic convergence between the extracted token clusters (\textit{taken-clusters}), meaning between the clipped categorical sub-dimensions of each initial category. We hypothesize this phenomenon of partial categorical confluence for the following reason: if categorical attention drives the extraction of a categorical sub-dimension from a precursor category in layer $n-1$, the coordination of this synthetic process with categorical phasing will mechanically tend to induce a clipping of semantically related sub-dimensions among different precursor categories. This is due to the mathematical construction of the aggregation function in the form $\sum(w_{i,j} x_{i,j}) + b$: for a given initial token, its resulting activation in the target neuron will be stronger (and thus, this token will be extracted and become a \textit{taken-token}) if it is simultaneously activated within multiple precursor categories—that is, if the $\sum$-effect, categorical phasing, operates. Categorical phasing, therefore, inherently results in partial semantic convergence.

To illustrate this concept, let us revisit the example from Graph 1:

\begin{itemize}
    \item Categorical sub-dimension $\alpha_1$ \textit{"speculative claims"} includes, among others, the tokens: \texttt{[claim], [allegedly], [claimed], [allege], [infamous], [notorious], [rumor].}
    \item Categorical sub-dimension $\alpha_2$ \textit{"legal allegations"} includes, among others, the tokens: \texttt{[claim], [allegedly], [claimed], [allege], [accusation], [charges].}
\end{itemize}

These two sub-dimensions share the common tokens \texttt{[claim], [allegedly], [claimed], [allege]}, thereby mechanically exhibiting a relative semantic convergence.

\subsection{Categorical Activational Dispersion}

A second conceptual and methodological approach in our investigation of the phenomenon of categorical restructuring is as follows.

What effect does the coactivity of synthetic categorical attention and categorical priming have on the activational aspect of artificial categorical restructuring during the transition from precursor categories to their associated category in a subsequent layer? Here, we hypothesize an effect of \textit{categorical activational dispersion}, meaning that a taken-token cluster, extracted from a precursor category, does not correspond to a continuous segment of activation values for these tokens within the implicated precursor neuron. In other words, a clipped categorical sub-dimension does not delineate a homogeneous activational segment (i.e., the taken-tokens within a given taken-cluster do not have similar activation values within the corresponding precursor neuron). Put differently, categorical restructuring is accompanied by \textit{activational restructuring}: the structure or topology of taken-token activations in precursor neurons is not preserved.

Why hypothesize such categorical activational dispersion? At first glance, it may seem counterintuitive. The phenomenon of categorical priming would suggest that tokens strongly activating a precursor neuron should also strongly activate its associated target neuron and therefore be present in the extracted taken-cluster. However, this reasoning overlooks the significant impact of the categorical phasing process. In other words, categorical priming alone is insufficient to necessarily produce strong activations in target neurons and thus to extract tokens. For a token to become a taken-token, it must not only strongly activate its precursor neuron (categorical priming) but also, statistically, be subject to categorical phasing (with another precursor neuron). However, these two synthetic cognitive phenomena are theoretically independent. Consequently, two tokens with similar activation levels (even high ones) within a precursor neuron are not necessarily both subject to categorical phasing: statistically, one may undergo the process while the other does not. As a result, the first will become a taken-token while the other, insufficiently activated in the target neuron, will not.

Furthermore, for similar reasons, we also hypothesize activational dispersion among taken-tokens at the level of the target neuron. That is, tokens within the same taken-cluster do not necessarily have similar activation levels within this target neuron. Therefore, a given categorical sub-dimension associated with a target category does not delineate a determined activational zone in this target neuron. This occurs, among other reasons, because two categorical sub-dimensions tend to share certain tokens in common (cf. the phenomenon of synthetic partial categorical confluence we hypothesized earlier), but this tendency is only partial. Thus, two taken-clusters associated with the same target neuron will likely share some tokens while differing in others. The shared tokens will undergo categorical phasing, while the distinct ones will not. Due to the mathematical construction of the aggregation function, this leads to discontinuities in activation values in the target neuron.

We further hypothesize this phenomenon of activational dispersion at the target neuron level because it aligns with previously observed properties of categorical discontinuity among successive core-tokens regarding their activation levels (postulating that successive core-tokens exhibit particularly low cosine similarities) and categorical inhomogeneity in mono-activation patterns among core-tokens (suggesting that core-tokens with similar activation levels are not necessarily the most categorically similar) in prior studies \cite{Pichat2024a}. Finally, this phenomenon of activational dispersion among taken-tokens at the target neuron level implies that categorical restructuring at the target neuron level is not necessarily accompanied by an isomorphic structuring of the activation space within this target neuron.

\subsection{Categorical Distancing}

We now present our third and final conceptual and methodological approach in operationalizing our research on categorical restructuring.

If the coactivity of the three mathematico-cognitive factors (categorical priming, attention, and phasing) involved in clipping indeed produces categorical restructuring, then it follows \textit{de jure} that the categorical segmentation system of a neural layer $n+1$ differs from that of its preceding layer $n$. In other words, the type of categorical segmentation employed by a target neuron differs from the type of segmentation of each of its precursor neurons. 

We therefore hypothesize the following property: \textit{synthetic categorical distancing} between a category carried by a neuron in layer $n+1$ and the categories vectorized by its predecessor neurons in layer $n$. More precisely, we will empirically investigate this characteristic of synthetic cognition by measuring the semantic distance between the categorical clusters of core-tokens of a target neuron and the categorical clusters of core-tokens of each of its precursor neurons with which it has a strong connection weight. This semantic distance will be measured within the observational reference frame constituted by the embeddings of GPT2-XL.

\section{Methodology}

\subsection{Methodological Framework}

To methodologically situate our present exploratory study, we provide here a brief overview of various explainability techniques. These techniques attempt, with varying degrees of cognitive acuity, to elucidate the informational content or processes of artificial neural networks, whether structured in terms of neural layers, groups of layers, or entire networks.

Explainability research adopting a broad cognitive perspective focuses on examining variations between inputs and outputs to clarify the relationship between input data and the results of a language model. Among these methods, gradient-based approaches assess the impact of each input feature by studying the partial derivatives associated with each input dimension \cite{Enguehard2023}. Input features can be analyzed through different aspects such as traits \cite{Danilevsky2020}, token importance \cite{Enguehard2023}, or attention weights \cite{Barkan2021}. Additionally, example-based approaches explore how outputs change with varying inputs, observing the effects of slight input modifications \cite{Wang2022} or transformations such as deletion, negation, shuffling, or masking of inputs \cite{Atanasova2020, Wu2020, Treviso2023}. Other studies focus on conceptual mapping of inputs to estimate their contribution to observed results \cite{Captum2022}.

Explainability methods with greater cognitive acuity focus on the model's intermediate internal states rather than just the final output, scrutinizing the partial outputs or internal states of neurons or neuron groups. In this context, some studies analyze and linearly decompose the activation of a neuron in a given layer concerning its inputs (neurons, attention heads, or tokens) in the previous layer \cite{Voita2021}. Other techniques aim to simplify activation functions to facilitate their interpretation \cite{Wang2022}. Some methods, leveraging the model’s lexicon, focus on extracting encoded knowledge by projecting connections and intermediate representations through a correspondence matrix \cite{Dar2023, Geva2023}. Finally, certain strategies rely on statistical analyses of neuronal activation in response to a dataset \cite{Bills2023, Mousi2023, Durrani2022, Wang2022, Dai2022}. Our current research falls precisely within this latter category of methods.

\subsection{Methodological Approach}

In preliminary research \cite{Pichat2024a, Pichat2024b, Pichat2024c, Pichat2024d, Pichat2024e}, we decided to examine the transformer model GPT developed by OpenAI, focusing specifically on the GPT-2XL version. This selection was motivated by the fact that GPT-2XL offers sufficient complexity for analyzing advanced artificial cognitive phenomena while being less complex than GPT-4 or its current multimodal version, GPT-4o. Another key factor in our decision was that, in 2023, OpenAI provided, through the publication of Bills et al. \cite{Bills2023}, comprehensive information on the model’s parameters and neuronal activation values, which are essential for our research.

To simplify our investigation, we concentrated our analysis on the first two perceptron layers of GPT-2XL, each comprising 6,400 neurons, for a total of 12,800 artificial neurons. Regarding tokens and their activation values in these neurons, we chose to examine, for each neuron, the 100 tokens with the highest average activation values, which we termed \textit{core-tokens}. Furthermore, in the context of this study on categorical restructuring—and again for simplification purposes—we focused, for each target neuron in layer 1, solely on its precursor neurons in layer 0 with the highest positive connection weights (up to a maximum of 10 precursor neurons per target neuron).

To assess semantic similarity between tokens, we opted to measure cosine similarity within the GPT-2XL embeddings database. We deliberately avoided using the GPT-4 embedding base, despite its superior performance, to circumvent the methodological limitation highlighted by Bills et al. \cite{Bills2023} and Bricken \cite{Bricken2023}, which warns against associating artificial cognitive systems that do not rely on the same embedding system—that is, on a different categorical segmentation framework.

\subsection{Statistical Options}

For our statistical studies, we utilized Python's SciPy libraries, in accordance with the statistical recommendations of Howell \cite{Howell2024} and Beaufils \cite{Beaufils1996}.

To assess the normality of our data—an essential characteristic for conducting parametric tests—we followed a two-phase approach. First, we performed inferential statistical tests, including: the Shapiro-Wilk test, which is effective for small datasets; the Lilliefors test, relevant when the parameters of the normal distribution are unknown and estimated from the data; the Kolmogorov-Smirnov test, suited for large datasets; and the Jarque-Bera test, which evaluates skewness and kurtosis in large samples. We then complemented this approach with descriptive statistics, such as skewness (for asymmetry) and kurtosis (for the degree of flattening), along with visual techniques like the QQ-plot to compare observed data to a theoretical normal distribution. To verify the homogeneity of variances between groups, we applied Bartlett’s test (sensitive to deviations from normality) as well as Levene’s test (less affected by such deviations) as a complementary measure.

The results, partially presented in this article, indicate an approximate normality of our data. Consequently, our statistical analyses primarily rely on non-parametric approaches, specifically:

\begin{itemize}
    \item The Kruskal-Wallis test, which explores the relationship between a categorical variable defining multiple independent groups and an ordinal variable. This test was applied by ranking our numerical data on token neuronal activation and, in accordance with its application conditions, for groups of at least five observations.
    \item The univariate chi-square ($\chi^2$) goodness-of-fit test, considering its requirements regarding theoretical and observed frequencies to avoid the need for alternatives for small sample sizes, such as Fisher’s exact test or Monte Carlo methods.
    \item The binomial test, ensuring the binarity of the involved variables, as well as the independence of trials and the equality of probability distributions across categories.
\end{itemize}

\section{Presentation of Our Results}

\subsection{Partial Categorical Confluence}

As previously indicated, our first approach to studying the synthetic phenomenon of categorical restructuring focuses on the effect produced by the coactivity of synthetic categorical attention and categorical phasing on this restructuring, specifically during the transition from precursor categories to their associated category in a subsequent layer. More precisely, we aim to better understand how this restructuring manifests at the level of the clipped categorical sub-dimensions of each precursor category.

Within this framework, we hypothesized the existence of a phenomenon in synthetic cognition: \textit{partial categorical confluence}. That is, for a target neuron (in layer 1 in our study), a relative semantic convergence between the extracted token clusters (\textit{taken-clusters}) from its strongest precursor neurons (in terms of connection weight); in other words, between the clipped categorical sub-dimensions of each initial category.

From a methodological perspective, our procedure was as follows. For each target neuron in perceptron layer 1, we focused on its 10 strongest precursor neurons in layer 0 (based on attention-weighted connection strength) and, more specifically, on the 10 clipped \textit{taken-clusters} from these precursors. We then compared each of these \textit{taken-clusters} against the others, considering only cases where both compared clusters contained at least six \textit{taken-tokens} (the minimum cardinality threshold for all \textit{taken-clusters} in this study, ensuring statistical applicability of the Kruskal-Wallis test). For each of these effective comparisons, the semantic distance between the involved tokens was evaluated in terms of cosine similarity using the observational reference frame constituted by GPT2-XL embeddings, according to the formula (excluding duplicates):

\[
m = \text{mean}(\cos(\text{tokens-cluster}_x, \text{tokens-cluster}_y))
\]

Table 1 presents an observed mean semantic distance of .45 between the associated \textit{taken-clusters}, indicating a relative semantic convergence among these clusters. Among the 9,463 effective comparisons, 72\% exhibit a cosine similarity below .5 (noting that cosine similarity ranges from 0 to 1); this finding is statistically significant ($p(\chi^2) < .0001$) with a chi-square goodness-of-fit test based on an equiprobability hypothesis.

These results are consistent with our hypothesis of \textit{partial categorical confluence}, suggesting that categorical phasing generates an intersection effect between categories while still allowing the independent influence of categorical attention in producing a categorical complementation effect. In other words, \textit{taken-clusters} extracted from precursor categories tend to converge semantically, yet this convergence does not equate to categorical identity (which would occur if \textit{taken-clusters} were identical, implying phasing-driven extraction without complementary differentiation). This conclusion, however, remains specific to the semantic observational reference frame provided by GPT2-XL embeddings.

\begin{table}[H]
    \centering
    \renewcommand{\arraystretch}{1}
    \small
    \rowcolors{1}{gray!20}{gray!40}
    \begin{tabular}{|c|c|}
        \hline
        N (effective crossings) & 9463 \\
        \hline
        Mean (Mean (m)) & .453 \\
        \hline
        \% of (Mean (m) < .5) & 72.241 \\
        \hline
        $p(\chi^2)$ of (Mean (m) < .5) & 6.78E-213 \\
        \hline
    \end{tabular}
    \captionsetup{justification=centering, format=plain}
    \caption*{\textit{Table n$^{\circ}$1: Cosine similarity statistics related to taken-clusters between layers 0 and 1 of GPT2-XL.}}

    \label{tab:table1}
\end{table}

\subsection{Categorical Activational Dispersion}

As discussed in the presentation of our research problem, our second axis of investigation into the synthetic phenomenon of categorical restructuring concerns the impact of the coactivity of categorical attention and priming on the activational aspect of this restructuring. We hypothesized an effect of \textit{categorical activational dispersion}, suggesting that a \textit{taken-token} cluster extracted from a precursor category does not correspond to a continuous segment of activation values for these tokens within the involved precursor neuron. In other words, an extracted categorical sub-dimension does not delineate a homogeneous activational zone (i.e., a segment of similar activation values).

Our methodological approach was as follows. For each \textit{taken-cluster}, we compared the mean activation distance (on the precursor neuron) among its constituent tokens to the first quartile of activation distances among the 100 \textit{core-tokens} (on that same precursor neuron). The first quartile represents the lower range of these distances. This was computed using the formula:

\[
    \begin{split}
        d = \text{mean}(&\lvert \text{activation}_{\text{taken-token}_x} - \text{activation}_{\text{taken-token}_y} \rvert) \\
        &- Q1 (\lvert \text{activation}_{\text{core-token}_n} - \text{activation}_{\text{core-token}_m} \rvert)
    \end{split}
\]

Positive values of $d$ indicate a lack of proximity between the activation values of \textit{taken-tokens}. This analysis was again restricted to (i) only the 10 strongest precursor neurons in terms of connection weight for each associated target neuron, and (ii) only \textit{taken-clusters} containing at least six tokens.

Table 2, based on 9,007 calculated distances, reports a mean distance value $d$ of .33. Additionally, 88.45\% of cases exhibit a positive distance value, a trend that is highly significant ($p(\chi^2) < .0001$). These results support our hypothesis of \textit{activational dispersion} and suggest that categorical restructuring, as precursor categories transition to their associated categories in the subsequent layer, is accompanied by an activational restructuring: activational token segments in a precursor neuron do not define the categorical segments of \textit{taken-clusters} that are clipped from the categories associated with these precursor neurons during restructuring.

\begin{table}[H]
    \centering
    \renewcommand{\arraystretch}{1}
    \small
    \rowcolors{1}{gray!20}{gray!40}
    \begin{tabular}{|c|c|}
        \hline
        N & 9007 \\
        \hline
        Mean (Mean (d)) & .331 \\
        \hline
        \% of (Mean(d) > 0) & 88.448 \\
        \hline
        $p(\chi^2)$ of (Mean(d) > 0) & 1.47E-14 \\
        \hline
    \end{tabular}
    \captionsetup{justification=centering, format=plain}
    \caption*{\textit{Table n$^{\circ}$2: Statistics on activation distance comparisons between taken-tokens and core-tokens (activations in layer 0 of GPT2-XL).}}

    \label{tab:table2}
\end{table}

The results presented in Table 3, established using the same methodology, confirm our hypothesis regarding \textit{activational dispersion} at the level of the target neuron. Specifically, the data indicate that \textit{taken-tokens} do not form activationally cohesive clusters within their respective target neurons. The calculated mean distance value is .43, with 88.4\% of cases exhibiting a positive $d$ value, a result that is highly significant ($p(\chi^2) < .0001$). These findings further support our hypothesis of \textit{activational dispersion} at the level of target neurons: the clipped categorical sub-dimensions, extracted from the precursor neurons' categories, do not delineate specific activational zones within their associated target neurons.

\begin{table}[H]
    \centering
    \renewcommand{\arraystretch}{1}
    \small
    \rowcolors{1}{gray!20}{gray!40}
    \begin{tabular}{|c|c|}
        \hline
        N & 9007 \\
        \hline
        Mean (Mean (d)) & .432 \\
        \hline
        \% of (Mean(d) > 0) & 88.385 \\
        \hline
        $p(\chi^2)$ of (Mean(d) > 0) & 1.63E-14 \\
        \hline
    \end{tabular}
    \captionsetup{justification=centering, format=plain}
    \caption*{\textit{Table n$^{\circ}$3: Statistics on activation distance comparisons between taken-tokens and core-tokens (activations in layer 1 of GPT2-XL).}}

    \label{tab:table3}
\end{table}

\subsection{Categorical Distancing}

Our final investigative angle on the synthetic phenomenon of categorical restructuring concerns the existence of \textit{categorical distancing} between the categorical segmentation represented by a target neuron and those represented by its strongest precursor neurons (in terms of connection weights). More specifically, we examine whether a difference exists between the categorical segmentation of a target neuron's category clusters and those of its precursor neurons. We investigate this hypothesis by evaluating the semantic distance between the categorical clusters of \textit{core-tokens} in a target neuron and those in each of its strongly connected precursor neurons. As in previous analyses, this semantic distance is measured within the observational reference frame provided by GPT2-XL embeddings.

From a methodological standpoint, it is important to emphasize that the categorical clusters analyzed here are not absolute segmentations but are relative to the operationalization method chosen for their generation. In this study, we employed \textit{prompt engineering} with GPT-4o to produce these clusters, maintaining a fixed number of five clusters per neuron. The categorical clusters of each target neuron's category were then compared to the categorical clusters of its precursor neurons’ categories. This comparison was again restricted to (i) only the 10 strongest precursor neurons (in layer 0) in terms of connection weight for each target neuron (in layer 1) and (ii) only categorical clusters containing at least six tokens. For each comparison between a precursor categorical cluster $x$ in layer 0 and a target categorical cluster $y$ in layer 1, we computed the semantic proximity between tokens in $x$ and $y$, as well as the internal proximity between tokens in $x$. This was calculated using the following formula:

\[
d = \text{mean}(\cos(\text{tokens}_x, \text{tokens}_y)) - \text{mean}(\cos(\text{tokens}_x))
\]

A negative value of $d$ indicates a significant semantic distance between target and precursor categorical clusters, supporting the effect of categorical distancing we are investigating.

Table 4 presents the results obtained from 4,692 target neurons in layer 1, each having at least one precursor neuron (in layer 0) with at least one categorical cluster containing at least six tokens. A total of 138,367 distances $d$ were computed, with a mean value of -.14, and an overwhelmingly high percentage (99.83\%) of cases where $d$ is negative. This trend is statistically significant ($p(\chi^2) < .0001$). Additionally, a substantial proportion (82.27\%) of cases showed a statistically significant distance under the Kruskal-Wallis test, again highly significant ($p(\chi^2) < .0001$).

\begin{table}[H]
    \centering
    \renewcommand{\arraystretch}{1}
    \small
    \rowcolors{1}{gray!20}{gray!40}
    \begin{tabular}{|c|c|}
        \hline
        $N_d$ & 138367 \\
        \hline
        Mean (Mean (d)) & -.142 \\
        \hline
        \% of (Mean(d) < 0) & 99.829 \\
        \hline
        $p(\chi^2)$ of (Mean(d) < 0) & 2.15E-23 \\
        \hline
        \% of ($p_{KW}$ < .05) & 82.274 \\
        \hline
        $p(\chi^2)$ of ($p_{KW}$ < .05) & 1.08E-10 \\
        \hline
    \end{tabular}
    \captionsetup{justification=centering, format=plain}
    \caption*{\textit{Table n$^{\circ}$4: Statistics on semantic distance comparisons between categorical clusters of target neurons (layer 1) and categorical clusters of their associated source neurons (layer 0).}}

    \label{tab:table4}
\end{table}

To complement this analysis, we performed an alternative evaluation of the semantic distance between target and precursor categorical clusters. Specifically, we counted the number $n$ of common tokens for each cluster comparison and computed the index:

\[
d' = n - \frac{n_x}{10}
\]

where $n_x$ represents the number of tokens in a given precursor categorical cluster $x$. A negative value of $d'$ thus indicates a relatively low proportion of common tokens.

Table 5 reveals a low average number (.35) of common tokens between a target neuron’s categorical cluster and each of the categorical clusters of its associated precursor neurons. Moreover, the relative mean number of common tokens is negative (-1.3), with an extremely high and significant percentage (97.81\%) of cases where $d'$ is negative. Furthermore, a large percentage (94.59\%) of cases show significance under a binomial test for $d'$ negativity, with this percentage itself being highly significant ($p(\chi^2) < .0001$).

\begin{table}[H]
    \centering
    \renewcommand{\arraystretch}{1}
    \small
    \rowcolors{1}{gray!20}{gray!40}
    \begin{tabular}{|c|c|}
        \hline
        $N_d$ & 138367 \\
        \hline
        Mean (Mean (n)) & .349 \\
        \hline
        Mean (Mean (d’)) & -1.298 \\
        \hline
        \% of (Mean(d’) < 0) & 97.811 \\
        \hline
        $p(\chi^2)$ of (Mean(d’) < 0) & 1.15E-21 \\
        \hline
        \% of ($p_b$ (d’<0) < .05) & 94.589 \\
        \hline
        $p(\chi^2)$ of ($p_b$ (d’<0) < .05) & 4.76E-19 \\
        \hline
    \end{tabular}
    \captionsetup{justification=centering, format=plain}
    \caption*{\textit{Table n$^{\circ}$5: Statistics on the count of common tokens between categorical target clusters (layer 1) and their corresponding source clusters (layer 0).}}

    \label{tab:table5}
\end{table}

These various results are compatible with our postulate of categorical distancing, which relates to the fact that categorical restructuring, occurring during the transition from the categorical segmentation of precursor neurons (layer 0) to the categorical segmentation of their corresponding target neurons (with high connection weights), manifests as a semantic gap between the synthetic categories carried by these respective source and target corollary neurons. This holds, at least when using the GPT2-XL embeddings as a semantic observation framework and comparing the categorical subgroups (categorical clusters) segmented by the formal neurons. Categorical restructuring, during the transition from one neural layer to another, indeed defines a new system for categorically partitioning the world of tokens. This is, of course, the primary function of successive synthetic neural layers, aimed at overcoming the initial categorical limitations of the input embeddings—embeddings that are not sufficiently efficient for performing the required tasks and for which the neural network has been specifically trained.

\section{Qualitative Illustration of Our Results}

In addition to our previous quantitative analyses, we now illustrate the main synthetic cognitive processes that we have mobilized in our investigation. This aims to qualitatively deepen our empirical understanding of how artificial categorical restructuring operates.

\subsection{Synthetic Categorical Complementation}

As a reminder, synthetic categorical attention, or the w-effect, refers to the fact that, during its categorical restructuring activity from layer \( n \) to layer \( n+1 \), a target neuron, due to its inherent aggregation function, will focus more on specific subordinate categories to form its own specific category. This results in a process of categorical complementation, which consists, genetically, in "bringing" to the extension (of tokens) of this target category a singular categorical sub-dimension extracted from a precursor category—a sub-dimension that is semantically distinct from and complementary to other categorical sub-dimensions extracted from the remaining precursor categories.

Graph 2 presents an example of categorical complementation generated by synthetic categorical attention (case of target neuron no. 1 in layer 1 of GPT2-XL). The target neuron (in layer 1), associated with the category \textit{"Cultural \& social contexts"}, extracts here, from its precursor neurons in layer 0, through complementation, three categorical sub-dimensions that are semantically distinct and complementary to one another:

\begin{itemize}
    \item \textit{\(\alpha_1\) "Latin American \& Spanish names \& places"}, extracted from precursor category no. 1778 \textit{"Cultural geography"}, containing, among others, the tokens: \texttt{[Argentine]}, \texttt{[Luis]}, \texttt{[Juan]}, \texttt{[Puerto]}, \texttt{[Nicarag]}.
    \item \textit{\(\alpha_2\) "Employment \& roles"}, clipped from precursor category no. 1657 \textit{"Categorical data"}, involving notably the tokens: \texttt{[transfer]}, \texttt{[incub]}, \texttt{[stint]}.
    \item \textit{\(\alpha_3\) "Football teams and roles"}, abstracted from precursor category no. 5065 \textit{"Sports entity names"}, encompassing various tokens such as: \texttt{[goalkeeper]}, \texttt{[relegation]}, \texttt{[Cardiff]}, \texttt{[Southampton]}.
\end{itemize}

\begin{figure}[H]
    \centering
    \hspace*{-.2\textwidth} 
    \includegraphics[width=1.4\textwidth]{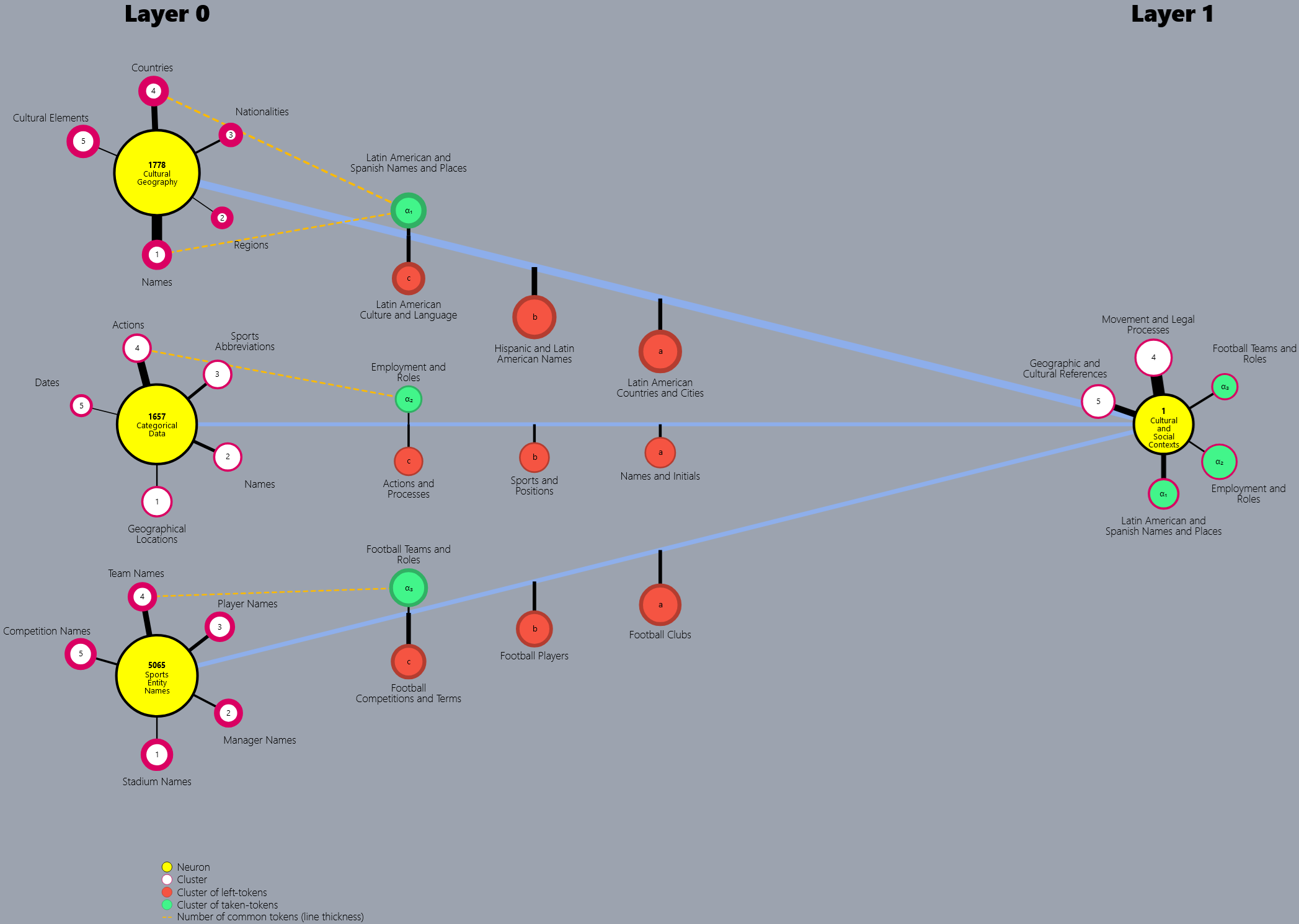}
    \captionsetup{justification=centering, font=small}
    \caption*{\textit{Graph n$^{\circ}$2: Case of categorical complementation generated by synthetic categorical attention (target neuron n°1 in layer 1 of GPT2-XL).}}

    \label{fig:graph2}
\end{figure}

\subsection{Synthetic Categorical Phasing}

As previously mentioned, synthetic categorical phasing, or the \(\Sigma\)-effect, relates to the fact that, during its categorical restructuring activity from layer \( n \) to layer \( n+1 \), a target neuron, given its particular aggregation function, extracts from several of its precursor categories categorical sub-dimensions that are semantically correlated with one another. This manifests as a process of categorical intersection, in which the same tokens, resonating categorically, are jointly extracted from the source categories and thus simultaneously present in the extension of the extracted sub-dimensions.

Graph 3 presents an example of categorical intersection produced by synthetic categorical phasing (case of target neuron no. 121 in layer 1 of GPT2-XL). The target neuron (in layer 1), associated with the category \textit{"Unverified assertions"}, extracts here, from its precursor neurons in layer 0, through intersection, three categorical sub-dimensions that are significantly semantically linked to one another:

\begin{itemize}
    \item \textit{\(\alpha_1\) "Speculative claims"}, extracted from precursor category no. 6356 \textit{"Document elements"}.
    \item \textit{\(\alpha_2\) "Legal allegations"}, derived from precursor category no. 3721 \textit{"Legal concepts"}.
    \item \textit{\(\alpha_3\) "Speculative claims"}, originating from precursor category no. 5207 \textit{"Linguistic elements"}.
\end{itemize}

These categorical sub-dimensions, as illustrated with a few non-exhaustive examples in Table 6, correspond to token-clusters containing a series of shared tokens.

\begin{figure}[H]
    \centering
    \hspace*{-.2\textwidth} 
    \includegraphics[width=1.4\textwidth]{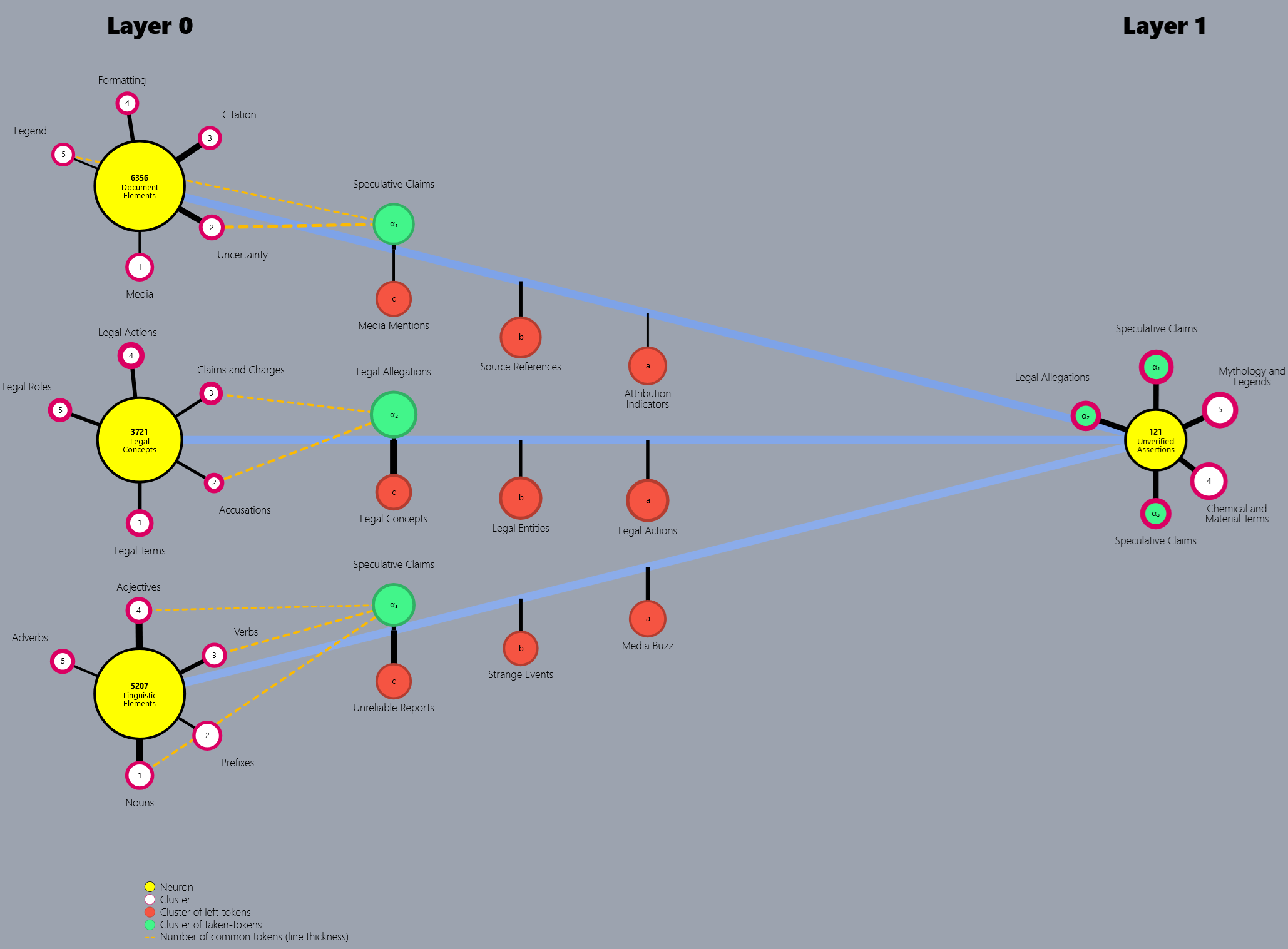}
    \captionsetup{justification=centering, font=small}
    \caption*{\textit{Graph n$^{\circ}$3: Case of categorical intersection generated by synthetic categorical phasing (target neuron n°121 in layer 1 of GPT2-XL).}}

    \label{fig:graph3}
\end{figure}

\begin{table}[H]
    \centering
    \renewcommand{\arraystretch}{1}
    \small
    \rowcolors{1}{gray!20}{gray!40}
    \begin{tabular}{|c|c|c|c|}
        \hline
        & $\alpha_1$ & $\alpha_2$ & $\alpha_3$ \\
        \hline
        \texttt{[claim]} & X & X & X \\
        \hline
        \texttt{[allegedly]} & X & X & X \\
        \hline
        \texttt{[purported]} & X & & X\\
        \hline
        \texttt{[reportedly]} & X & & X \\
        \hline
        \texttt{[alleges]} & & X & X\\
        \hline
        \texttt{[alleged]} & & X & X\\
        \hline
    \end{tabular}
    \captionsetup{justification=centering, format=plain}
    \caption*{\textit{Table n$^{\circ}$6: Identical tokens between taken-clusters $\alpha_1$, $\alpha_2$, $\alpha_3$ extracted respectively from predecessor categories n$^{\circ}$6356, 3721, and 5207 in layer 0, by the target neuron n°121.}}

    \label{tab:table6}
\end{table}

\subsection{Synthetic Categorical Clipping}

Categorical attention and phasing, along with categorical priming, are the synthetic processes (driven by neural aggregation functions) that shape categorical clipping during categorical restructuring. Clipping, as previously mentioned, is the mechanism in artificial cognition that actively constructs a categorical form, distinguishing it from a categorical background (which is also constructed). This clipping process generates the categorical sub-dimension (materialized as a token-cluster) that is uniquely abstracted from each source category.

Graph 4 presents a case of semantic categorical clipping: the token-tokens constituting the categorical sub-dimension \(\alpha_1\), extracted as \textit{"Enhanced modifiers"} from the category \textit{"Linguistic elements"} of precursor neuron no. 2945 (layer 0), are semantically homogeneous (at least when using human semantics as a reference framework). These tokens include (in part): \texttt{[Hyper]}, \texttt{[Super]}, \texttt{[turbo]}. The categorical sub-dimension \(\alpha_1\) is actively constructed rather than merely passively identified from an intrinsically preexisting semantic subgroup; indeed, this sub-dimension is semantically distinct from other possible categorical clusters of this neuron, such as: \textit{"Proper nouns"}, \textit{"Abbreviations"}, \textit{"Capitalized words"}, \textit{"Compound words"}, and \textit{"Prefixes"}.

Furthermore, this sub-dimension—clipped categorical form—is also semantically distinct from the unextracted categorical background of the source category, which can be segmented into the following three left-clusters:

\begin{itemize}
    \item \textit{Left-cluster a "Prefixes"}, composed, among others, of the tokens: \texttt{[altern]}, \texttt{[aux]}, \texttt{[counter]}, \texttt{[subst]}.
    \item \textit{Left-cluster b "Superlative terms"}, including notably the tokens: \texttt{[superflu]}, \texttt{[superhuman]}, \texttt{[superpower]}, \texttt{[superior]}, \texttt{[supers]}.
    \item \textit{Left-cluster c "Variants \& modifiers"}, composed, for example, of the tokens: \texttt{[mini]}, \texttt{[doub]}, \texttt{[dual]}, \texttt{[extra]}, \texttt{[triple]}, \texttt{[sup]}.
\end{itemize}

\begin{figure}[H]
    \centering
    \hspace*{-.2\textwidth} 
    \includegraphics[width=1.4\textwidth]{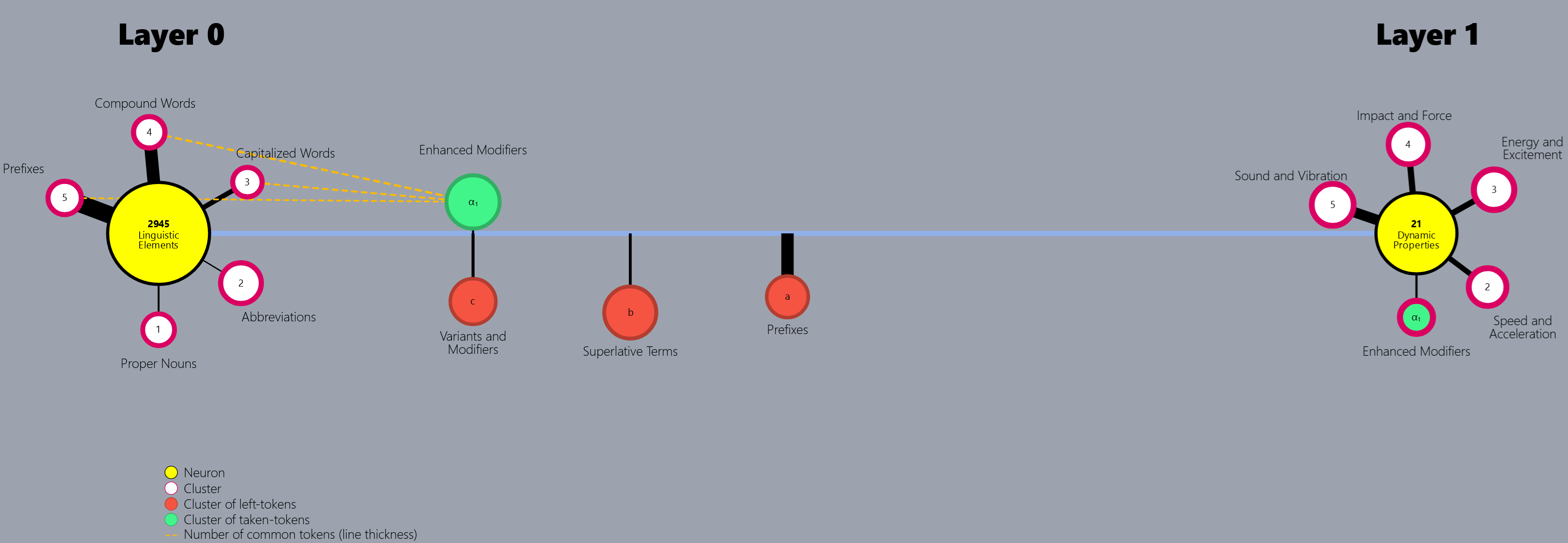}
    \captionsetup{justification=centering, font=small}
    \caption*{\textit{Graph n$^{\circ}$4: Case of synthetic categorical clipping of the semantic type (target neuron n°21 in layer 1 of GPT2-XL).}}

    \label{fig:graph4}
\end{figure}

Graph 5, on the other hand, allows us to observe a case of categorical clipping of a graphemic-phonological type:

\begin{itemize}
    \item The token-tokens constituting the categorical sub-dimension \(\alpha_1\), extracted as \textit{"Variations of be"} from the category \textit{"Lexical patterns"} of precursor neuron no. 3264 (layer 0), are homogeneous from a graphemic and/or phonological perspective. The tokens involved here include, for example: \texttt{[be]}, \texttt{[beau]}, \texttt{[bes]}, \texttt{[bei]}, \texttt{[beaver]}. The unique categorical sub-dimension \(\alpha_1\) is distinct from other possible categorical clusters of this source neuron, such as: \textit{"Prefixes"}, \textit{"Repeated tokens"}, \textit{"Complete words"}, \textit{"Capitalized tokens"}, and \textit{"Incomplete words"}.

    \item The token-tokens forming the clipped categorical sub-dimension \(\alpha_2\), \textit{"Vowel combinations"}, extracted from the category \textit{"Categorical nomenclature"} of precursor neuron no. 4664 (layer 0), are homogeneous from a graphemic and/or phonological perspective. Indeed, the tokens involved here include: \texttt{[AE]}, \texttt{[EA]}, \texttt{[EE]}, \texttt{[IE]}, \texttt{[EEE]}. This specific categorical sub-dimension \(\alpha_2\) is also distinct from other possible categorical clusters of this source neuron, such as: \textit{"Countries"}, \textit{"Religious terms"}, \textit{"Family members"}, \textit{"Abbreviations"}, and \textit{"Names"}.

    \item The token-tokens generating the categorical sub-dimension \(\alpha_3\), \textit{"Repeated vowels \& consonants"}, detached from the category \textit{"Human-related categories"} of precursor neuron no. 3555 (layer 0), converge at the graphemic and/or phonological level. These tokens include, among others: \texttt{[EE]}, \texttt{[oo]}. This specific categorical sub-dimension \(\alpha_3\) is differentiated from other possible categorical clusters of this source neuron, such as: \textit{"Names \& proper nouns"}, \textit{"Footwear \& movement"}, \textit{"Headwear \& accessories"}, \textit{"Body parts \& features"}, and \textit{"Containers \& structures"}.

    \item Finally, the token-tokens defining the categorical sub-dimension \(\alpha_4\), \textit{"Germanic sound patterns"}, clipped from the category \textit{"Linguistic elements"} of precursor neuron no. 5259 (layer 0), are coherent at the graphemic and/or phonological level. This sub-dimension contains the tokens: \texttt{[Die]}, \texttt{[Bei]}, and \texttt{[Lie]}. This distinct categorical sub-dimension \(\alpha_4\) is also different from other possible categorical clusters of this source neuron, such as: \textit{"Pronouns"}, \textit{"Names"}, \textit{"Abbreviations"}, \textit{"Prefixes"}, and \textit{"Conjunctions"}.
\end{itemize}

\begin{figure}[H]
    \centering
    \hspace*{-.2\textwidth} 
    \includegraphics[width=1.4\textwidth]{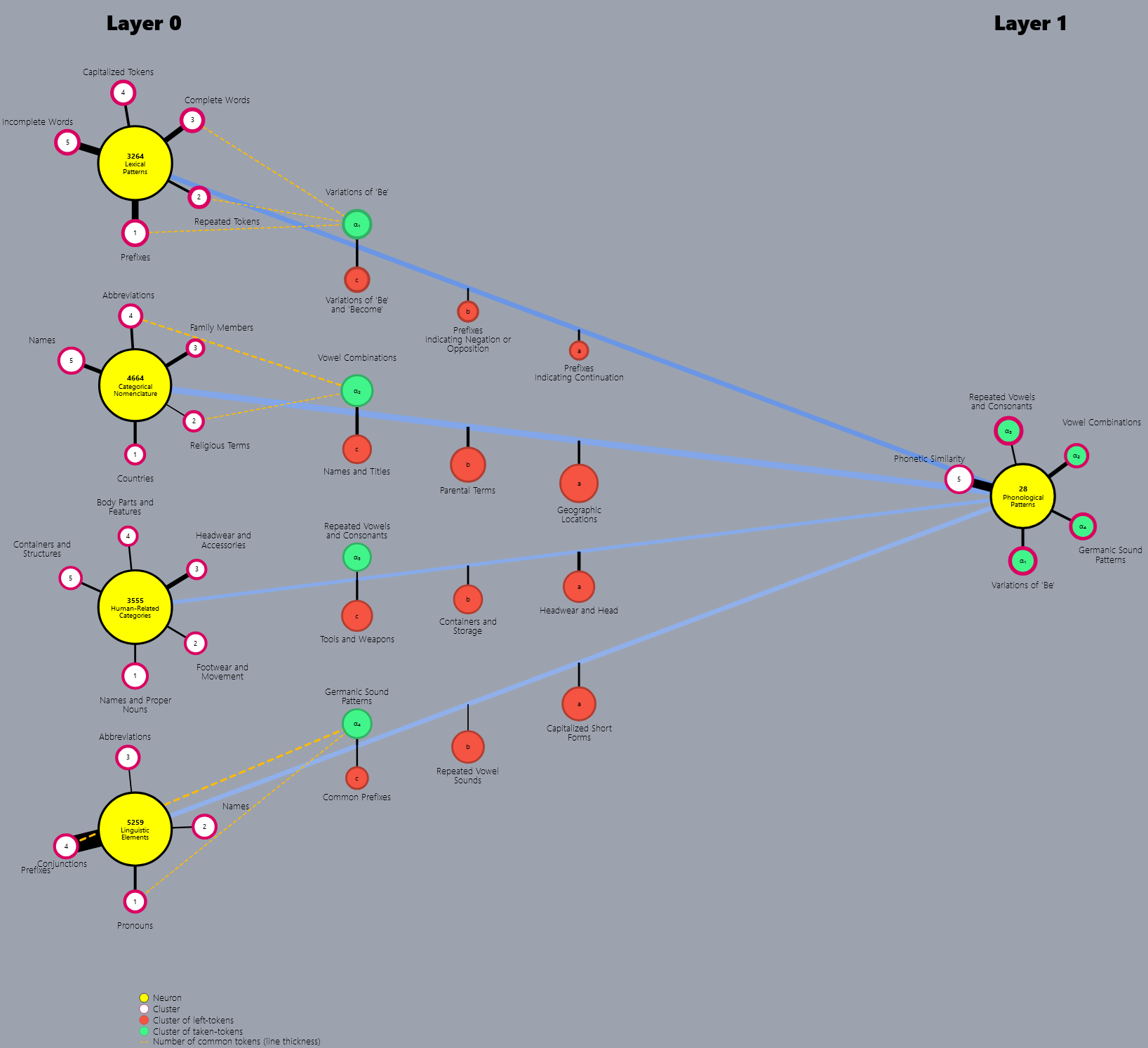}
    \captionsetup{justification=centering, font=small}
    \caption*{\textit{Graph n$^{\circ}$5: Case of synthetic categorical clipping of the graphemic-phonological type (target neuron n°28 in layer 1 of GPT2-XL).}}

    \label{fig:graph5}
\end{figure}

\subsection{Synthetic Categorical Restructuring}

The synthetic mathematical-cognitive factors of categorical priming, attention, and phasing shape the process of categorical clipping, which once again allows the category of a target neuron in layer \( n+1 \) to genetically form by extracting (and combining) from each of its precursor categories (in layer \( n \)) a singular categorical sub-dimension. These clipped categorical sub-dimensions become (potential) constitutive sub-dimensions of the newly formed superordinate category, thereby generating a new categorical segmentation system of the token world—an original system distinct from each of the involved precursor categories. This phenomenon is what we denote as the synthetic process of categorical restructuring.

Graph 6 provides an insightful illustration of the categorical restructuring process. From the first precursor neuron in layer 0, no. 6255, associated with the thought category \textit{"Current affairs"}, the categorical sub-dimension \(\alpha_1\) \textit{"Disasters \& threats"} is clipped. It contains, among others, the tokens: \texttt{[qaida]}, \texttt{[nightmare]}, \texttt{[malware]}, \texttt{[ransomware]}. One can easily observe how these tokens have been selectively extracted from at least four of the five categorical clusters that initially described the source category (these tokens being semantically linked to certain aspects of these categorical clusters): \textit{"Health"}, \textit{"Political Graphs / entities"}, \textit{"Technology"}, and \textit{"Criminal activities"}. This extracted sub-dimension, the first element of the categorical restructuring observed here, is semantically a genuine categorical construction, distinctly separate from the categorical background that was deemed irrelevant. This categorical background can be classified into three left-clusters:

\begin{itemize}
    \item \textit{Left-cluster a "Political Graphs \& entities"}, composed, among others, of the tokens: \texttt{[Maduro]}, \texttt{[Hezbollah]}, \texttt{[Chavez]}, \texttt{[Boko]}, \texttt{[Fidel]}, \texttt{[scientology]}.
    \item \textit{Left-cluster b "Natural \& environmental events"}, including the tokens: \texttt{[armageddon]}, \texttt{[tempest]}, \texttt{[typhoon]}, \texttt{[tornado]}, \texttt{[wildfire]}, \texttt{[irma]}.
    \item \textit{Left-cluster c "Health \& medical conditions"}, comprising notably the tokens: \texttt{[leukemia]}, \texttt{[herpes]}, \texttt{[pox]}, \texttt{[allergies]}, \texttt{[tumor]}, \texttt{[diabetes]}.
\end{itemize}

Still in Graph 6, we observe that another categorical sub-dimension, \(\alpha_2\) \textit{"Magical \& mystical elements"}, is extracted from precursor neuron no. 6040, belonging to the synthetic category \textit{"Symbolic elements"}. The extension of this sub-dimension includes, among others, the tokens: \texttt{[spiral]}, \texttt{[scourge]}, \texttt{[ginny]} (from Ginny Weasley, a character in \textit{"Harry Potter"}). These tokens were likely selectively extracted from at least four of the five categorical clusters that originally segmented the source category: \textit{"Spirituality"}, \textit{"Names"}, \textit{"Jewelry \& gems"}, and \textit{"Weapons"}. This second clipped sub-dimension, another part of the categorical restructuring described here, is again an original categorical construction, distinct from the disregarded categorical background. This categorical background can be decomposed into the following three left-clusters:

\begin{itemize}
    \item \textit{Left-cluster a "Spiritual \& mystical concepts"}, partially composed of the tokens: \texttt{[soul]}, \texttt{[virtue]}, \texttt{[grace]}, \texttt{[sacrament]}, \texttt{[gift]}, \texttt{[spirit]}, \texttt{[mystery]}.
    \item \textit{Left-cluster b "Weapons \& tools"}, including among others the tokens: \texttt{[spear]}, \texttt{[sword]}, \texttt{[dagger]}, \texttt{[lightsaber]}, \texttt{[scythe]}, \texttt{[baton]}.
    \item \textit{Left-cluster c "Colors \& gems"}, consisting notably of the tokens: \texttt{[sparkle]}, \texttt{[jewel]}, \texttt{[gem]}, \texttt{[pearl]}, \texttt{[ruby]}.
\end{itemize}

As part of the categorical restructuring detailed in Graph 6, we see that the clipped categorical sub-dimensions \(\alpha_1\) \textit{"Disasters \& threats"} and \(\alpha_2\) \textit{"Magical \& mystical elements"}, among other potential clipped sub-dimensions from various precursor categories in layer 0, may constitute categorical clusters of the newly associated category in layer 1. This new synthetic superordinate category, named \textit{"Catastrophic phenomena"}, can be described using three additional categorical clusters:

\begin{itemize}
    \item \textit{Cluster 3 "Natural disasters"}, composed, among others, of the tokens: \texttt{[calam]}, \texttt{[catastrophic]}, \texttt{[pandemonium]}, \texttt{[katrina]}, \texttt{[turmoil]}, \texttt{[havoc]}.
    \item \textit{Cluster 4 "Aggressive entities"}, including notably the tokens: \texttt{[malicious]}, \texttt{[ferocious]}, \texttt{[insurgents]}, \texttt{[demonic]}, \texttt{[taliban]}, \texttt{[devils]}, \texttt{[jihad]}, \texttt{[rebel]}.
    \item \textit{Cluster 5 "Emotional states"}, involving, for example, the tokens: \texttt{[stirred]}, \texttt{[messed]}, \texttt{[raging]}, \texttt{[mad]}.
\end{itemize}

Thus, before our very eyes, so to speak, a categorical restructuring has taken place through the process of categorical clipping, driven by the synthetic factors of categorical priming, attention, and phasing. From an initial categorical segmentation in layer 0, notably around the categories \textit{"Current affairs"} and \textit{"Symbolic elements"}, emerges in layer 1 a new, original synthetic category, \textit{"Catastrophic phenomena"}, whose progressive genesis we have just partially followed.

\begin{figure}[H]
    \centering
    \hspace*{-.2\textwidth} 
    \includegraphics[width=1.4\textwidth]{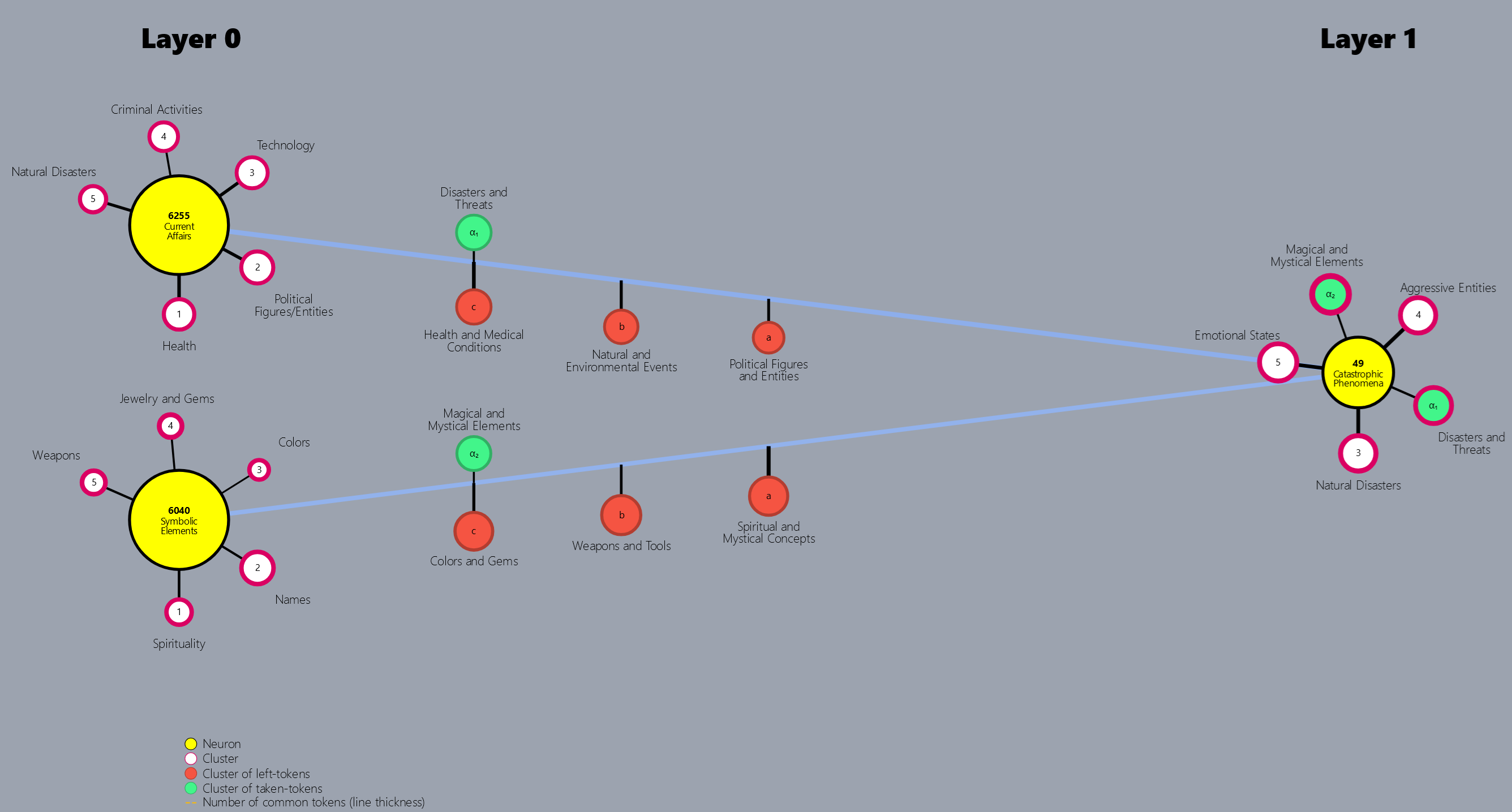}
    \captionsetup{justification=centering, font=small}
    \caption*{\textit{Graph n$^{\circ}$6: Case of synthetic categorical restructuring (target neuron n°49 in layer 1 of GPT2-XL).}}

    \label{fig:graph6}
\end{figure}
Our \href{https://neuron-viewer.neocognition.org/1/1}{genetic neuron viewer}, associated with this study, allows for the visualization of the synthetic categorical restructuring phenomenon occurring during the transition from perceptron layer 0 to 1 in GPT2-XL.

\section{Discussion}

\subsection{Synthesis of Our Empirical Results}

We have investigated the phenomenon of \textit{synthetic categorical restructuring}, which refers to the construction, at each new neural layer $n+1$, of new synthetic thought categories that are more efficient for segmenting the token space, thereby enabling the language model to perform its designated tasks. This restructuring process involves combining antecedent categories to create new ones, in a manner analogous to Piaget’s concept of \textit{reflective abstraction} \cite{Pichat2024e}.

We sought to understand how this categorical restructuring is achieved through the process of \textit{categorical clipping} \cite{Pichat2024e}, which consists of the following operations at each neuron of layer $n+1$:
\begin{itemize}
    \item Extraction of categorical forms (i.e., clipped categorical sub-dimensions) from the categories carried by its precursor neurons in layer $n$.
    \item Omission of non-retained categorical background deemed irrelevant.
\end{itemize}

Additionally, we aimed to analyze how this restructuring and clipping process is shaped by the activity and coactivity of three factors governing categorical segmentation \cite{Pichat2024d}, implemented through the neural aggregation function:
\begin{itemize}
    \item \textbf{Categorical priming (x-effect):} The stronger a token's activation within a category in layer $n$, the more likely it is to be extracted and become part of the extension of a category in layer $n+1$ to which it is strongly linked.
    \item \textbf{Categorical attention (w-effect):} The stronger the connection weight between a category in layer $n+1$ and a category in layer $n$, the more likely tokens from the precursor category will be extracted and incorporated into the target category.
    \item \textbf{Categorical phasing ($\Sigma$-effect):} The stronger a token’s simultaneous activation across multiple categories in layer $n$, the more likely it is to be included in the extension of a strongly connected category in layer $n+1$.
\end{itemize}

We have focused particularly on the synthetic phenomenon of categorical attention in our study of categorical restructuring, insofar as:

\begin{itemize}
    \item It enables and catalyzes the two other synthetic phenomena, categorical priming and phasing, which can only operate significantly when neurons in successive layers are strongly connected.
    \item Neural attentional connection weights are the regularities constructed and learned from the world of tokens to which the language model is exposed during its training phase; these regularities are fabricated to allow the language model to construct an interpretative order during its interaction with this world.
\end{itemize}

First, we demonstrated that the coactivity of categorical attention and categorical phasing generates the synthetic phenomenon of \textit{partial categorical confluence}, a fundamental property of categorical restructuring. Specifically, the clipped categorical sub-dimensions of categories represented by layer $n$ neurons, when linked to a strongly connected neuron in layer $n+1$, tend to converge within a semantic space to contribute to the formation of the target neuron’s category. This occurs because, for tokens to be extracted from a category in layer $n$ (i.e., to become \textit{taken-tokens} forming an extracted sub-dimension), their activation function $\Sigma(w_{i,j} x_{i,j}) + b$ at the target neuron must be strong. This implies a high degree of categorical phasing ($\Sigma$-effect), meaning that these tokens must be present in the extracted sub-dimensions (i.e., \textit{taken-clusters}) of multiple precursor categories. Mechanically, this leads to semantic convergence among the involved categorical sub-dimensions within the restructuring process.

Second, we demonstrated that the interaction of categorical attention and categorical priming induces a synthetic process of \textit{categorical dispersion}. This means that an extracted categorical sub-dimension of a category represented by a neuron in layer $n$ does not correspond to a continuous activation segment (i.e., tokens with similar activation values) within the activation space of the precursor neuron. In other words, activational token segments in a precursor neuron do not define the categorical segments of \textit{taken-clusters} that will be clipped during categorical restructuring. We also observed this dispersion effect in the activation space of the target neuron. This dispersion is primarily due to the significant impact of categorical phasing during clipping: categorical priming alone does not necessarily result in sufficiently strong activation in the target neurons to extract tokens as part of a clipped sub-dimension. For a token in layer $n$ to become a \textit{taken-token}, it must not only strongly activate its precursor neuron (categorical priming) but also be involved in categorical phasing with one or more additional precursor neurons. These two processes are \textit{a priori} independent. Consequently, two tokens with similar activation levels (even high ones) within a precursor neuron are not necessarily both subject to categorical phasing—statistically, one may be, while the other is not.

Third, we demonstrated that synthetic categorical restructuring involves a phenomenon of \textit{categorical distancing}: the categorical segmentation of objects (tokens) in categories represented by layer $n$ neurons is semantically different from the categorical segmentation in their strongly connected target neurons in layer $n+1$. This effect is particularly evident when semantic distance is measured within the observational reference frame of the model’s initial embeddings. Categorical distancing is, of course, the fundamental function of successive synthetic neural layers, as it allows the model to overcome the limitations of its initial embeddings—embeddings that are insufficiently efficient for the tasks the model is designed to perform.

\subsection{Functional Interpretation of Our Results}

As pointed out by von Glaserfeld \cite{vonGlaserfeld2002} and Varela \cite{Varela1984, Varela1988}, the activity of a cognitive system aims to extract functional regularities within the informational flow of its interaction experience with the external world to which it is exposed—the world of tokens in the context of language models.

This extraction is goal-oriented: it serves the efficiency of this intelligent system (efficiency informed by the feedback administered to this AI system during its deep learning phase). This extraction is not passive but active: it does not identify intrinsic and pre-given properties of the external world but rather constructs, based on the mathematical and architectural parameters specific to the synthetic system, as well as its training data and received feedback, functional invariants within the singularity of this system’s experience of the external world of tokens with which it interacts. This extraction of regularities, at least at the level of perceptron-type neurons, results in the learning of attentional weights, constituting the encryption key—a theorem-in-action in the sense of Vergnaud \cite{Vergnaud2009, Vergnaud2016, Vergnaud2020}—which determines how to combine, through a dynamic of reflective abstraction analogous to that of Piaget \cite{Piaget1974}, the synthetic thought categories (concepts-in-action in the sense of Vergnaud again \cite{Vergnaud2009}) of a layer at level \( n \) to form those of the next layer, making the latter even more discriminative and thus functional.

Categorical restructuring is the hallmark of this goal-oriented, active, and attentional extraction. It is the result of synthetic categorical clipping, produced by the cognitive-mathematical factors of categorical priming, attention, and phasing. It manifests through partial categorical convergence, activation dispersion, and categorical distancing, all of which are also determined by these cognitive-mathematical factors.

This synthetic categorical restructuring is the activity that, equivalently to what Varela \cite{Varela1984, Varela1988} describes regarding intelligent living systems, realizes the structural coupling that constitutes the formation of an intelligent system. Structural coupling occurs throughout the deep learning history of this synthetic system, progressively collecting and selecting (in the Latin etymological sense of intelligence) the appropriate attentional weights through which it is relevant to combine and shape its categorical interconnection system for analyzing its functional experience of interaction with the world of tokens. And it is this categorical restructuring that ultimately enables this intelligent system to "attach itself to a pre-existing world of meaning," as Varela would say, that of tokens and human meanings, with which it must coordinate to conceptualize (in the sense of Vergnaud ) and act effectively.

\section{Conclusion}

Language models internally structure their representation of the world by progressively segmenting and restructuring their own analytical modalities for processing objects from this world—tokens and their relationships. The categorical restructuring phenomenology that we have sought to elucidate in this study reveals how, at each new neural layer within a perceptron network, key sub-dimensions of the previous layer’s thought categories are clipped and combined to form new, increasingly effective synthetic categories. These newly structured categories enable the artificial system to better segment and process its experience of words and concepts from the external world, allowing it to functionally couple with these elements. 

In an upcoming study, currently in progress, we aim to investigate whether the categorical segmentations generated through synthetic restructuring are systematically associated with specific segmentation patterns in the activation spaces of the artificial neurons that carry these categorical structures.

\section*{Acknowledgments}

The authors would like to thank Madeleine Pichat for her careful review of this article and Stéphane Fadda (Sorbonne Center for Artificial Intelligence) for the valuable operational insights he provides to the Neocognition team.The authors would like to thank Albert Yefimov (Sberbank \& National University of Sciences \& Technologies of Moscow) for the stimulating philosophical and epistemological reflections on AI shared with him.

\end{document}